# Comparative Analysis of Novel View Synthesis and Photogrammetry for 3D Forest Stand Reconstruction and extraction of individual tree parameters

Guoji Tian[a,b,c], Chongcheng Chen[a,b,c], Hongyu Huang[a,b,c,*]

a     National Engineering Research Center of Geospatial Information Technology, Fuzhou University, Fuzhou 350108, China; 225527044@fzu.edu.cn (G.T.); chencc@fzu.edu.cn (C.C.)
b     Key Laboratory of Spatial Data Mining and Information Sharing of Ministry of Education, Fuzhou University, Fuzhou 350108, China
c     The Academy of Digital China (Fujian), Fuzhou 350108, China
\*     Correspondence: hhy1@fzu.edu.cn; Tel.: +86-13328269460

**Abstract**：Accurate and efficient 3D reconstruction of trees is crucial for forest resources assessments and management. Close-Range Photogrammetry (CRP) is widely used in 3D model reconstruction of forest scenes. However, in practical forestry applications, challenges such as low reconstruction efficiency and poor reconstruction quality persist. Recently, Novel View Synthesis (NVS) technology such as Neural Radiance Fields (NeRF) and 3D Gaussian Splatting (3DGS) has shown great potential in the 3D reconstruction of plants using some limited number of images. However, existing research typically focuses on small plants in orchards or individual trees. It remains uncertain whether this technology can be effectively applied in larger, more complex stands or forest scenes. In this study, we collected sequential images of forest stand plots with varying levels of complexity using different imaging devices. We then performed dense reconstruction of forest stand using NeRF and 3DGS methods. The resulting point cloud models were compared with those obtained through photogrammetric reconstruction and laser scanning methods. The results show that compared to photogrammetric methods, NVS methods have a significant advantage in reconstruction efficiency. Photogrammetric method is less suited to more complex forest stands, resulting in tree point cloud models with issues such as excessive canopy noise, wrongfully reconstructed trees with, duplicated trunks and canopies. In contrast, NeRF is better adapted to more complex forest stands, especially in reconstructing canopy regions. However, it can lead to reconstruction errors in the ground area when the input views are limited. The 3DGS method has a relatively poor ability to generate dense point clouds, resulting in models with low point density, particularly with sparse points in the trunk areas, which affects the accuracy of the diameter at breast height (DBH) estimation. Tree height information can be extracted from the point clouds reconstructed by all three methods, with NeRF achieving the highest accuracy. However, the accuracy of DBH extracted from photogrammetric point clouds is still higher than that from NeRF point clouds. These findings suggest that NVS methods based on sequential images have significant potential for 3D reconstruction of forest stands, providing further technical support for complex forest resource inventory and visualization tasks.

**Keywords:** 3D reconstruction；close-range photogrammetry (CRP)；neural radiance field (NeRF)；3D Gaussian Splatting（3DGS）；photogrammetry；deep learning；forest stand

## 1. Introduction

Forests are a crucial component of the Earth's ecosystem, fulfilling numerous ecological functions that impact soil conservation, climate regulation, wildlife habitats, noise and pollution reduction, and much more(Kankare et al., 2014). Comprehensive and accurate forest surveys provide essential baseline data needed for maximizing the multiple benefits of forests, maintaining ecological balance, strengthening forest resource management, and ensuring their rational use. The efficiency and accuracy of single-tree parameter measurements directly affects the precision and timeliness of forest resource surveys(Liang et al., 2022). Therefore, the ability to automatically, accurately, and quickly obtain individual tree parameters has become important in forest resource surveys.

The traditional method for obtaining tree parameters in forest stands involves manual measurements, such as using calipers or diameter tapes to measure the diameter at breast height (DBH) and hypsometers to measure tree height (TH), followed by recording and organizing the results. This approach is time-consuming and labor-intensive, and it may be prone to errors caused by human factors. 3D reconstruction technology can convert forest scenes into 3D digital models, enabling automatic extraction of individual tree parameters from these models. This overcomes some of the limitations of traditional survey methods. For example, many researchers use Terrestrial Laser Scanning (TLS) to obtain high-density 3D point cloud models of forest scenes, from which parameters such as TH and DBH can be derived for individual trees(Chen et al., 2019a; Chen et al., 2019b; Schneider et al., 2019). Due to the limited scanning angles of TLS and occlusion effects, it is challenging to capture canopy information for taller parts of trees. Therefore, laser scanners mounted on drones are used to collect canopy point cloud data. Jaskierniak et al. (2021) utilized an Airborne Laser Scanning (ALS) system to acquire point cloud data of structurally complex mixed forests and performed individual tree detection and canopy delineation. Liao et al. (2022) investigated the role of ALS data in improving the accuracy of tree volume estimation. Their study demonstrated that TH extracted from ALS data are more accurate than those measured manually with a telescoping pole. Combining ALS data with on-site DBH measurements effectively improves the accuracy of tree volume estimates. Although TLS and ALS can acquire large-area point cloud data in a short period, these devices are expensive and require high levels of technical expertise to operate. For example, some UAV operators need to obtain a permission to conduct flights. Additionally, these devices are heavy and cumbersome, which limits their use in complex forest environments. Alternatively, close-range photogrammetry (CRP) method has demonstrated significant potential in forest survey works(Chai et al., 2023; McGlade et al., 2022; Yan et al., 2024; Zhang et al., 2017). 3D point cloud was reconstructed from overlapping images. This allows for the acquisition of 3D model information of trees using only forest scene imagery which is easily available. Compared to TLS and ALS, CRP significantly reduces costs and operational complexity. The current mainstream method is Structure from Motion (SfM) (Schonberger and Frahm, 2016) + Multi-view Stereo (MVS) (Seitz et al., 2006), which involves feature detection, feature matching, and depth fusion between image pairs. SfM uses matching constraints and triangulation principles to obtain three-dimensional sparse point clouds and camera parameters, while MVS densifies the sparse point clouds. Kameyama and Sugiura (2020) used UAV to capture forest images under different conditions, employed the SfM method to create 3D models, and validated the measurement accuracy of tree height and volume. Bayati et al. (2021) used hand-held digital cameras and SfM-MVS to produce a 3D reconstruction of deciduous and uneven-aged forests and successfully extracted individual tree DBH with high accuracy. Xu et al. (2023) compared the accuracy of extracting forest structure parameters from SfM and Backpack LiDAR Scanning (BLS) point clouds. Their results showed that SfM point cloud models are well-suited for extracting DBH, but there is still a gap in the accuracy of TH extraction. This discrepancy depends on the quality of the point cloud model, which is influenced not only by the robustness of the algorithm but also by the quality of the acquired images and feature matching. In complex forest environments, there are often occlusions and varying lighting conditions between trees, which affect the quality of image data. Additionally, similar shape and texture features between trees pose challenges for feature point matching. To improve the quality of forest scene images, Zhu et al. (2021) compared Median Gaussian Filtering, Single-Scale Retinex, and Multi-Scale Retinex image enhancement algorithms to determine the most suitable algorithm for 3D reconstruction of forest scenes. Although these enhancements can improve the quality of forest scene reconstruction to some extent, there are still discrepancies in data accuracy compared to TLS, and there remains significant room for improvement in terms of time efficiency.

Recently, Novel View Synthesis (NVS) technology has become an active research topic in computer vision. Mildenhall et al. (2021) first introduced a new deep learning rendering method called Neural Radiance Fields (NeRF). NeRF implicitly renders complex static scenes in 3D using a fully connected deep network. Since then NeRF have drawn the attention of many researchers, leading to various improvements. Müller et al. (2022) introduced a hash mapping technique to enhance the sampling point positional encoding method in the original neural network, effectively accelerating network training. In terms of reconstruction accuracy, Barron et al. (2021) used conical frustums instead of ray sampling to address aliasing issues at different distances, while Wang et al. (2023) replaced the Multi-Layer Perceptron (MLP) with Signed Distance Functions (SDF) for geometric representation, achieving high-precision geometric reconstruction. These improvements have pushed research on NeRF into practical applications, providing high-quality 3D rendering perspectives for fields like autonomous driving (Cao et al., 2024) and 3D city modeling (Tancik et al., 2022). NeRF is not only capable of synthesizing novel view images but can also be used to reconstruct 3D models. Currently reported applications in 3D reconstruction include cultural heritage (Croce et al., 2023) and plants and trees. In the context of plant 3D reconstruction and phenotypic research, Hu et al. (2024) evaluated the application of NeRF in the 3D reconstruction of low-growing plants. The results demonstrated that NeRF introduces a new paradigm in plant phenotypic analysis, providing a powerful tool for 3D reconstruction. Zhang et al. (2024) proposed the NeRF-Ag model, which realized the three-dimensional reconstruction of orchard scenes and effectively improved the modeling accuracy and efficiency. In our previous research (Huang et al., 2024), we evaluated NeRF's ability to generate dense point clouds for individual trees of varying complexity, providing a novel example for NeRF-based single tree reconstruction. Nevertheless, NeRF still faces challenges in achieving high-resolution real-time rendering and efficient dynamic scene editing. 3D Gaussian Splatting (3DGS) (Kerbl et al., 2023) brought a technological breakthrough to the field. Unlike NeRF, 3DGS utilizes an explicit Gaussian point representation to precisely capture and present information about 3D scenes. In terms of optimization speed, rendering speed, and accuracy, 3DGS significantly outperforms the original NeRF technology, advancing NVS technology to a new level. 3DGS technology has become a transformative force driving innovations in related fields. Thanks to differentiable rendering optimization techniques based on 3D Gaussian point clouds, its research popularity continues to rise with notable achievements in areas such as novel view rendering (Ren et al., 2024) and dynamic scene reconstruction (Fan et al., 2024). However, the application of 3DGS to plants and forest scenes reconstruction and generation of dense point cloud models of tree have not yet been fully explored and evaluated.

Although research has applied NVS technology to the 3D reconstruction of complex-textured natural organisms, such as orchard plants and individual trees, there is still limited research on the application of NVS technology to large-scale forest stand with multiple trees. Therefore, this paper applies NeRF and 3DGS technologies to the 3D reconstruction of trees in forest stands to obtain dense 3D point cloud models. The generated point cloud models are then compared and evaluated against traditional photogrammetric methods using TLS point cloud as reference. The specific research objectives are as follows:

(1) Comparing the practical application of NVS techniques and traditional photogrammetric reconstruction methods in complex forest stands;

(2) Evaluating the ability of different NVS methods（one based on implicit neural networks：NeRF；another on explicit Gaussian point clouds：3DGS）in reconstructing trees and generating dense point clouds;

(3) Comparing tree parameters extracted from various 3D point cloud models and assessing whether NVS techniques can replace or supplement traditional photogrammetric methods, potentially becoming a new tool for forest scene reconstruction and forest resource surveys.

## 2. Materials and Method

### 2.1. Study Area

We selected two contrasting forest stand plots within the Qishan campus of Fuzhou University to conduct this experiment. Plot_1 is an irregularly elongated circular area with a topography that is higher in the center and lower around the edges. It is located in a relatively open area with minimal surrounding vegetative obstruction, facilitating image data collection. There are 45 golden rain trees (*Koelreuteria bipinnata 'integrifoliola'*) within the plot. During data collection, which occurred in the winter leaf-off period, the tree trunks and the canopies are clearly visible. Plot_2 is also an irregularly shaped area with a similar topography, being higher in the center and lower around the edges. One side of the plot is relatively open, while the other side is densely occupied with other trees, presenting challenges for image acquisition. The predominant tree species is the autumn maple tree (*Bischofia javanica Blume*), an evergreen broadleaf tree. The plot features 33 trees with high density, significant height variation, large canopy spread and dense foliage. Fig. 1 shows the morphology of the study plots. For the research objectives, we selected forest stand plots with different tree types and levels of complexity to study how these factors might affect the reconstruction results.

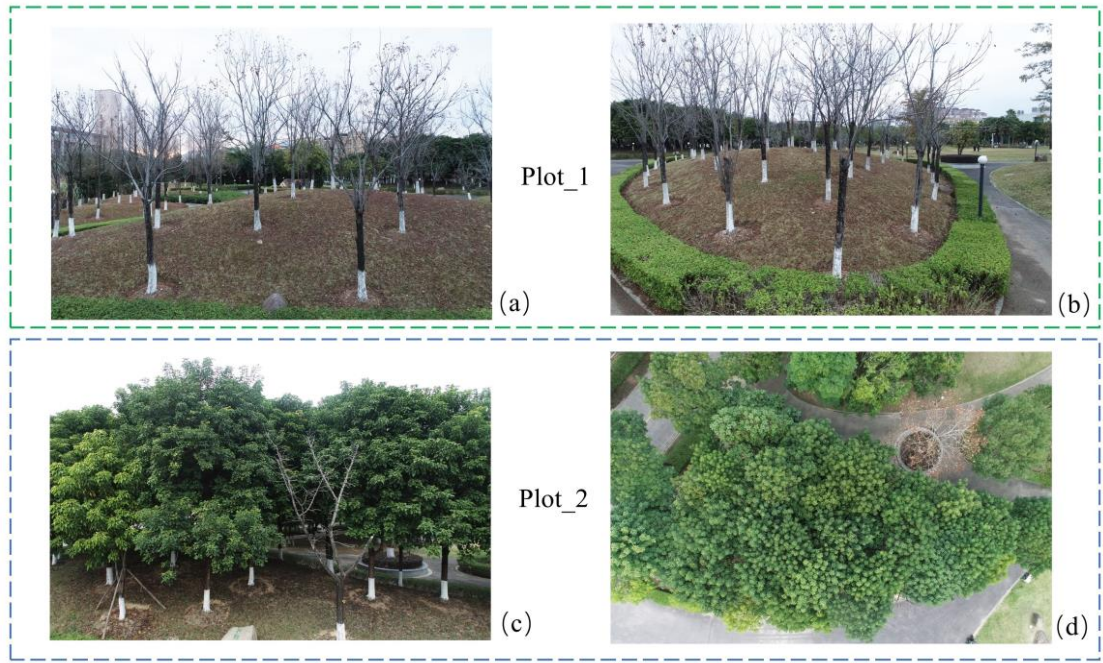

**Fig. 1.** The structures and shapes of two plots used in this study. Upper panel: (a, b) Plot_1 with leafless trees as observed from two views from mid-air; Lower panel: (c, d) Plot_2 with leafy trees viewed from mid-air and from overhead position.

*2.2. Research Method*

*2.2.1. Traditional Photogrammetric Reconstruction*

Photogrammetric reconstruction based on computer vision has become a relatively mature processing routine. Structure from Motion (SfM) and Multi-View Stereo (MVS) are the two main steps. SfM determines the spatial and geometric relationships of objects through camera motion. It starts with feature extraction algorithms to identify feature points from sequential images, followed by feature matching algorithms to match these points and eliminate those that do not meet certain geometric constraints. The final results of SfM are the external and internal parameters of the camera, as well as sparse feature points. MVS then uses the corresponding images and camera parameters obtained from the sparse reconstruction to generate a dense point cloud model through multi-view stereo vision. Specifically, from a pair or multiple pairs of calibrated images, the depth of each pixel in the scene is estimated through disparity computation, and a cost function is used to generate depth maps. Finally, these depth maps are fused to generate dense point clouds using voxel-based, point-based, or graph-based methods (Merrell et al., 2007). In this study, the photogrammetric reconstruction experiments were conducted using COLMAP (https://colmap.github.io/,V3.8), a widely used open-source SfM and MVS program in 3D reconstruction research. Additionally, both NeRF and 3DGS also rely on calibrated images, camera pose parameters, and sparse point clouds generated by COLMAP's SfM module as initial inputs for training.

*2.2.2. Neural Radiance Fields（NeRF）*

NeRF represents an entire scene using a multi-layer perceptron (MLP) as an implicit function. Its input consists of the position $X$ in 3D space and the viewing direction $d$, and its output is the RGB color $c$ and the density $\sigma$ at that point. The function is expressed as:

$$F_\theta : (X, d) \rightarrow (c, \sigma) \tag{1}$$

The formula includes $X = (x, y, z)$, which represents the position coordinates of a point in 3D space; $d = (\theta, \phi)$, which is the 2D camera viewing direction; and $c = (R, G, B)$, which is the color of the point. Since the color may vary under different viewing angles, it is related to both the viewing direction $d$ and the spatial position $X$.

For scenes represented by neural implicit functions, raster projection rendering is typically challenging because the surface information of the scene is not directly provided. Therefore, NeRF uses volumetric rendering to obtain RGB images from new viewpoints. Additionally, research by Rahaman et al. (2019) shows that MLP networks are more inclined to learn low-frequency signals in space, whereas natural scenes often contain complex geometry and significant color variations. Directly using positional coordinates as input makes it difficult to fit the high-frequency signals present in space. Therefore, NeRF employs positional encoding to map low-frequency positional information into a higher-dimensional space, making it easier for the network to fit the input space and learn high-frequency representations.

NeRFStudio (Tancik et al., 2023) is a modular PyTorch framework for NeRF development that integrates various NeRF methods, helping researchers to experience and learn these models more quickly. It provides a convenient web interface (viewer) to display training and rendering processes in real-time, and allows exporting rendering results as videos, point clouds, and mesh data. The default method in NeRFStudio is Nerfacto (Zhang et al., 2021), which leverages the strengths of several methods to enhance performance, ensuring both high accuracy and computational efficiency. We chose to conduct NeRF Forest stand reconstruction experiments within the NeRFStudio framework.

*2.2.3. 3D Gaussian Splatting（3DGS）*

3DGS defines the radiance field of a 3D scene on a discrete 3D Gaussian point cloud to achieve differentiable volume rendering. The technical process is shown in Fig. 2. In 3DGS, each point is represented as an independent 3D Gaussian distribution, which is mathematically expressed as:

$$G(x) = \exp\left(-\frac{1}{2}(x-\mu)^T \Sigma (x-\mu)\right) \quad (2)$$

Where $\mu$ and $\Sigma$ represent the mean and covariance matrix of the 3D Gaussian distribution, respectively. Each Gaussian point $P$ is also assigned an opacity $o$ and a color value $c$ to represent the radiance field of the 3D scene. Due to phenomena such as specular reflection, shadows, and occlusions, the color of the same object appears varying from different viewing angles. To model the characteristic of color value changes with viewing angles, 3DGS employs spherical harmonics (SHs).

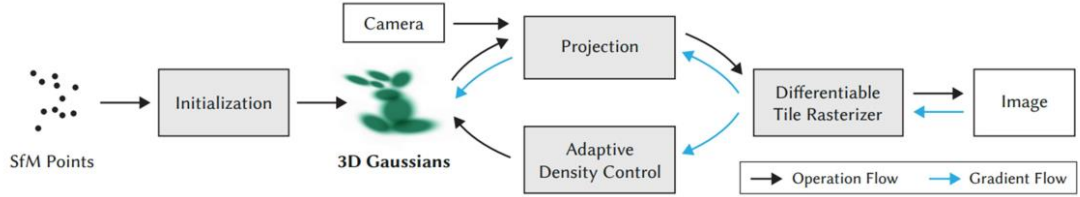

**Fig. 2.** Overview of 3D Gaussian Splatting workflow（from reference (Kerbl et al., 2023)）.

The rendering process of 3DGS, specifically in the differentiable rendering optimization process, uses the Object-Order approach. The basic idea is to use Elliptical Weighted Average (EWA) techniques (Zwicker et al., 2002) to calculate the projection of 3D Gaussian distributions and spherical harmonic functions onto a two-dimensional plane, based on the Gaussian points in three-dimensional space and the camera position. This is followed by the overlay and blending of Gaussians on the two-dimensional plane to ultimately synthesize the rendered image. In the backward phase, the loss is evaluated by calculating the difference between the rendered image and the real image, which is then used for backpropagation. The gradients from backpropagation are ultimately used to optimize the mean, covariance matrix, opacity, and spherical harmonic coefficients of the 3D Gaussian points.

Another key factor in the success of 3DGS is the adaptive control of the density of Gaussian points, as shown in Fig. 3. Based on the area occupied by the 2D Gaussian points on the plane, 3DGS further classifies the Gaussian points to be refined into two categories: Over-Reconstruction and Under-Reconstruction. For the over-reconstructed Gaussian points, 3DGS performs a cloning operation, while for the under-reconstructed points, it implements a splitting operation. This approach effectively increases the number of Gaussian points in the 3D scene, significantly improving rendering quality. In addition to generating new views, the result of 3DGS can also be exported in the form of mesh or point cloud.

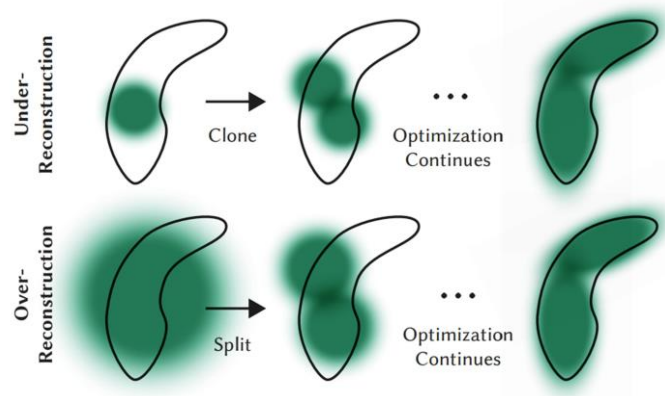

**Fig. 3.** 3D Gaussian splatting technique utilizes adaptive density control for densification （from reference (Kerbl et al., 2023)）.

*2.3. Data Acquisition and Processing*

*2.3.1. Data Acquisition*

We used multiple devices and methods for data collection. A smartphone camera was used to capture video footage of Plot_1 on the ground and individual images were later extracted from video frames. Additionally, a camera on DJI Phantom 4 UAV was employed to manually collect image data of both Plot_1 and Plot_2 from both ground and aerial perspectives. Fig. 4 shows the camera positions for Plot_1 and Plot_2 obtained after being processed with COLMAP. These consumer-grade cameras are not only inexpensive but also easy to operate, and are widely used by the general public. The image resolution and number of images are detailed in Table 1. Using a RIEGL VZ-400 terrestrial laser scanner (with parameters: accuracy of 3 mm, precision of 5 mm, laser beam divergence of 0.35 mrad, angular resolution of 0.04 degrees), multi-station scans were conducted on two sample plots to obtain ground truth reference point cloud data. The point clouds from multiple scans were then coarsely and finely registered in the RiSCanPro software (http://www.riegl.com/products/software-packages/riscan-pro/), with a registration accuracy of approximately 4 mm.

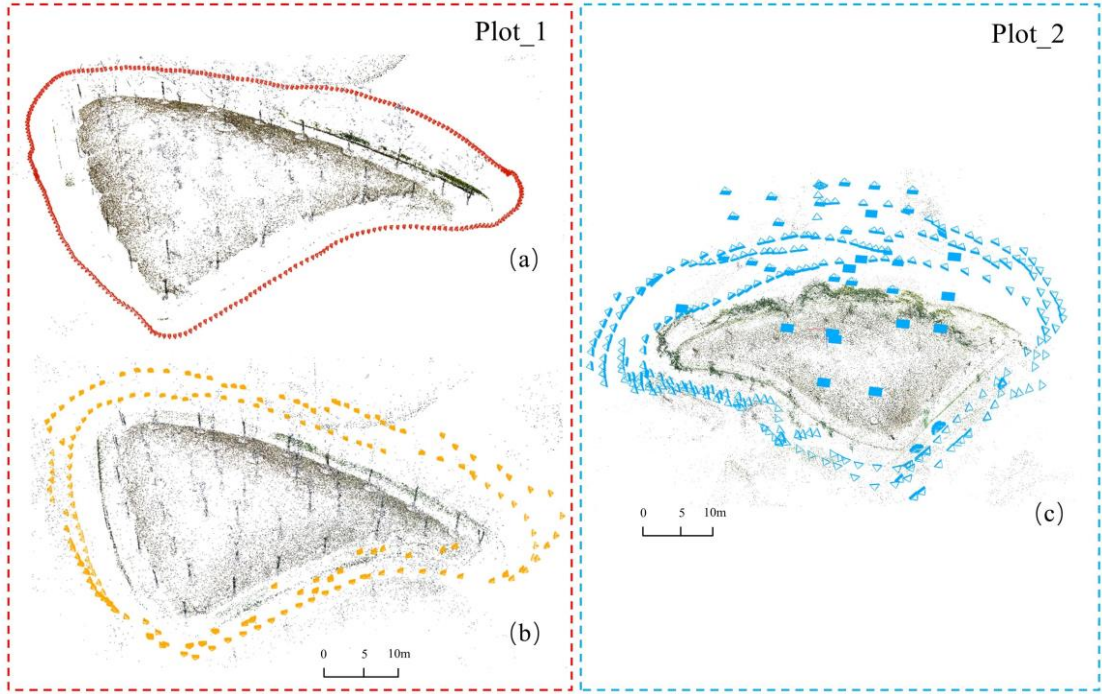

**Fig. 4.** Camera positions obtained using COLMAP for two plots. Left panel: (a) and (b), the red and yellow markers indicate the camera positions corresponding to each image, with (a) showing Plot_1 image taken with smartphone camera and (b) showing images taken with camera on UAV. Left panel: (c), the blue markers represent the camera positions for each image taken by the camera on UAV in Plot_2.

**Table 1.** Image data information for two forest stand plots.

| Image Dataset | Number of Images | Image Resolution |
| --- | --- | --- |
| Plot_1_Phone | 279 | 2160×3840 |
| Plot_1_UAV | 268 | 5472×3648 |
| Plot_2_UAV | 322 | 5472×3648 |

**Note:** The first part of the image dataset type name represents the plot name, and the second part describes the image acquisition sensor, including smartphone cameras (Phone) and drones (UAV). For example, Plot_1_Phone refers to data collected on Plot_1 using a smartphone camera.

*2.3.2. Data Processing*

The first step in photogrammetry, NeRF, and 3DGS reconstruction is the same SfM, which involves using the captured image data to obtain the corresponding camera poses and sparse point cloud representations. SfM results are used as input for MVS in photogrammetric processing. The entire photogrammetry reconstruction process is implemented using the open-source COLMAP. The NeRF method was implemented within the NeRFStudio. It utilized camera pose parameters and calibration images obtained from SfM for new view synthesis and 3D model creation. Once completed, the dense

3D point cloud model of the scene was exported. In the training of 3DGS, sparse point clouds obtained from SfM are used as initial values to construct an initial 3D Gaussian point cloud for rendering and optimization. The resulting densified and optimized scene point cloud is then exported Registrations with the TLS point cloud were performed to assign real-world scale information to the dense point cloud models generated by photogrammetry, NeRF, and 3DGS. Finally, we conducted a comprehensive evaluation and analysis by visually and quantitatively comparing the distribution of and tree parameters extracted from different point cloud models. Fig. 5 illustrates the workflow of this study. The cloud server system used for the reconstruction experiments (photogrammetry, NeRF, and 3DGS) was configured with Linux operating system, 12-core CPU, 24 GB RAM, NVIDIA GeForce RTX4090 (24 GB VRAM) GPU, deep learning framework of PyTorch 2.0 and CUDA 11.8.

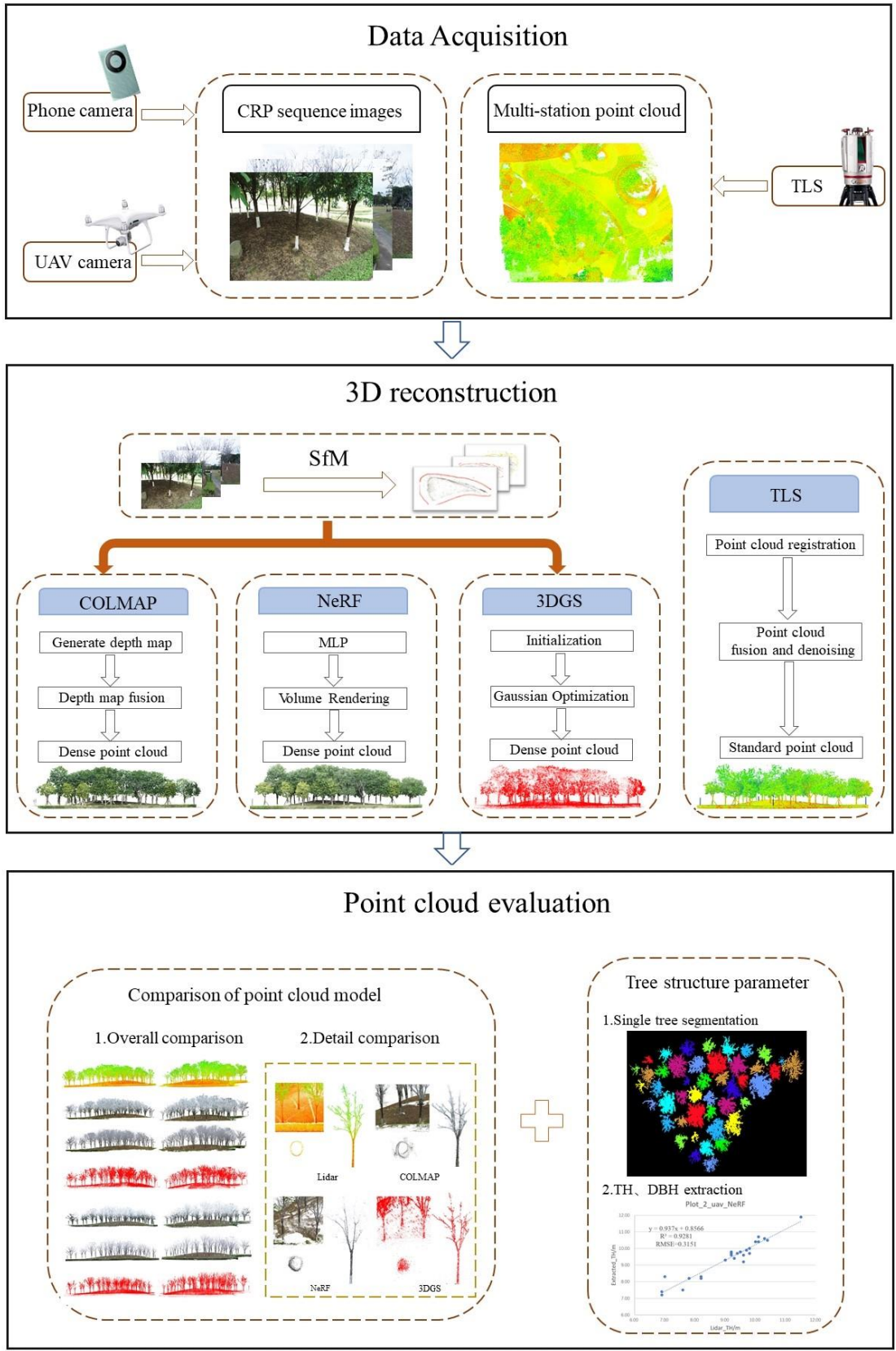

**Fig. 5.** Complete workflow of this study. Data Acquisition involves taking sequential images from various angles using different cameras, as well as laser scanning to obtain reference point cloud; during 3D reconstruction images were processed with COLMAP to obtain camera poses and sparse point clouds, which were further fed into three separate reconstruction methods: photogrammetry (COLMAP), NeRF, and 3DGS to generate dense points. Finally, these dense point clouds are registered and compared with standard point cloud obtained from TLS for a comprehensive evaluation of the point cloud models.

# 3. Results

*3.1 Reconstruction Efficiency Comparison*

Since SfM is a common step among the three reconstruction methods, the MVS dense reconstruction process in photogrammetry is equivalent to the training process in NeRF and 3DGS. We will compare the time required for dense point cloud generation in COLMAP with the time required for training NeRF for 30,000 epochs and 3DGS for 20,000 epochs. In NeRF and 3DGS, training is typically completed after a fixed number of epochs. However, we observe that 30,000 epochs for NeRF and 20,000 epochs for 3DGS are generally enough to achieve optimal reconstruction results. As shown in Table 2, COLMAP requires the most time to generate dense point clouds, while NeRF requires the least time. The time required for NeRF reconstruction ranges between 12-15 minutes, and the reconstruction time for 3DGS is approximately 1.3 times that of NeRF to realize similar visual effect. In contrast, COLMAP is 37 to 51 times slower than the other two reconstruction methods. Overall, the time required for the two NVS reconstruction technologies is significantly less than that for traditional photogrammetric reconstruction methods.

**Table 2.** Computation time (minutes) of dense reconstruction for different image datasets and dense reconstruction methods.

|        | Plot_1_Phone | Plot_1_UAV | Plot_2_UAV |
|--------|--------------|------------|------------|
| COLMAP | 544.292      | 724.495    | 453.834    |
| NeRF   | 15.0         | 14.0       | 12.0       |
| 3DGS   | 18.23        | 17.39      | 17.46      |

*3.2 Point Cloud Comparison*

We imported the point clouds obtained from COLMAP, NeRF, and 3DGS reconstructions into CloudCompare (https://www.cloudcompare.org/), registered them with the laser scanner point cloud, and performed visual comparisons. The number of point clouds in the plot as obtained from various methods are summarized in Table 3.

Among the three reconstruction methods, COLMAP generates the most point clouds, followed by NeRF, and 3DGS generates the fewest. The number of points generated by COLMAP is 4.4 to 20 times that of NeRF and 13 to 66 times that of 3DGS. Additionally, models reconstructed using images from UAV and phones show that UAV_COLMAP generates twice as many points as Phone_COLMAP, while the point counts for UAV models in NeRF and 3DGS are fewer compared to their corresponding Phone models.

Table 3. Number of points of the tree point cloud models.

| Tree ID | Model ID | Number of Point |
|---|---|---|
| Plot_1 | Plot_1_Lidar | 25,617,648 |
| | Plot_1_Phone_COLMAP | 20,200,476 |
| | Plot_1_Phone_NeRF | 4,548,307 |
| | Plot_1_Phone_3DGS | 1,555,984 |
| | Plot_1_UAV_COLMAP | 53,153,623 |
| | Plot_1_UAV_NeRF | 2,573,330 |
| | Plot_1_UAV_3DGS | 806,149 |
| Plot_2 | Plot_2_Lidar | 9,053,897 |
| | Plot_2_UAV_COLMAP | 55,861,268 |
| | Plot_2_UAV_NeRF | 5,465,952 |
| | Plot_2_UAV_3DGS | 831,164 |

**Note:** The first part of the Model ID name indicates the name of the plot, the middle part describes the image acquisition sensor, which can be a smartphone camera (Phone) or an unmanned aerial vehicle (UAV), and the last part indicates the image reconstruction method, which can be COLMAP, 3DGS, or NeRF.

Fig. 6 shows the overall spatial distribution of point cloud for Plot_1, seeing from two distinct perspectives. Clearly, Lidar as the reference offers the best quality. The other three reconstruction methods successfully reconstruct Plot_1 visually. COLMAP and NeRF models contain color information, while 3DGS does not capture real-world color information and is therefore shown in red. The COLMAP model contains the most ground and trees and has the highest number of points, but it has more noise on the tree canopy and branches. The Phone_NeRF model captures the trees fully, but the ground coverage is uneven. Compared to Phone_NeRF, the UAV_NeRF model has fewer point clouds but features a more even ground and less noise on the tree canopy. The 3DGS model has the fewest point clouds and appears sparser.

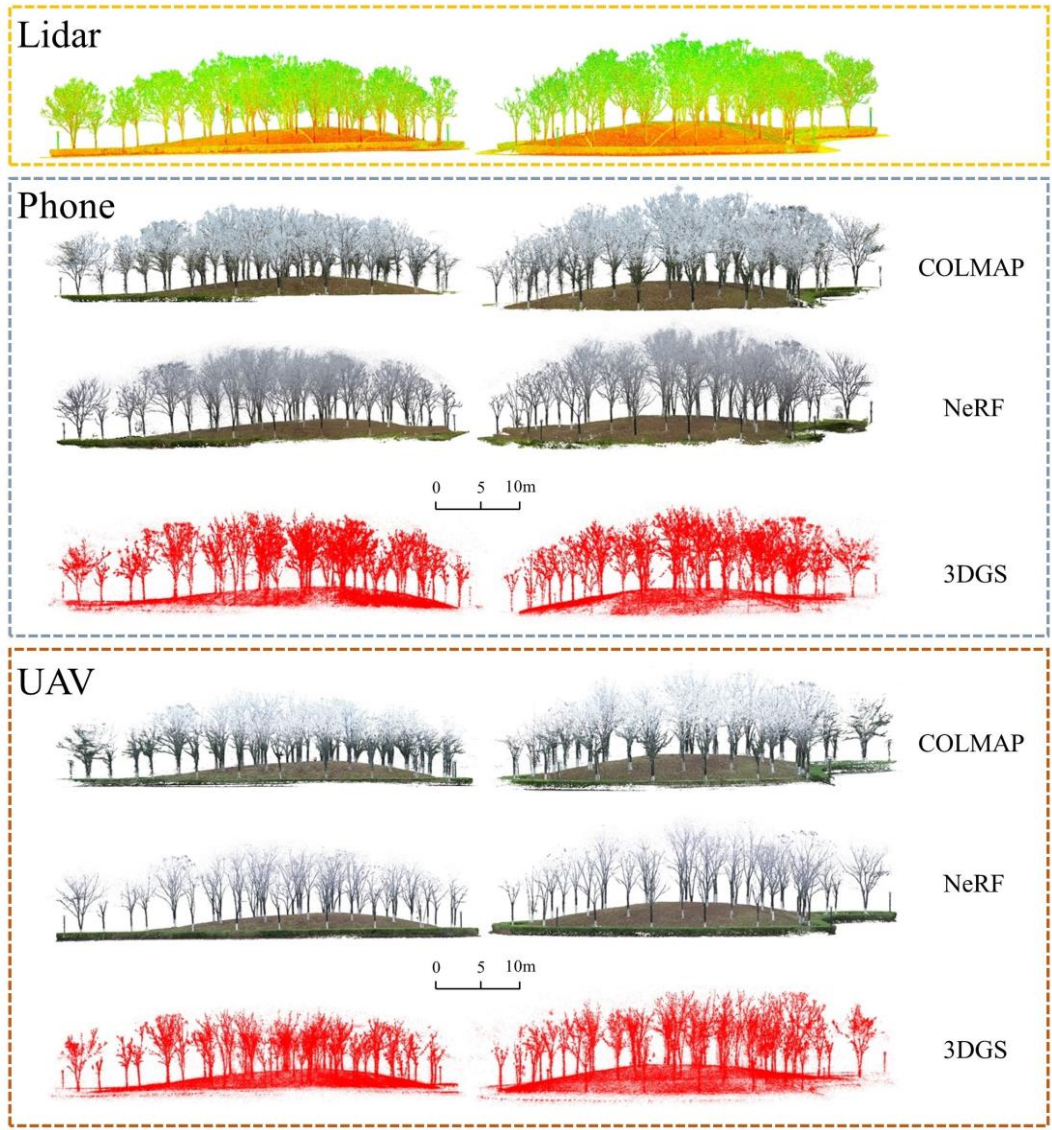

**Fig. 6.** Overall comparison of the reconstruction results for Plot_1 presented from two opposing viewing angles. TLS Lidar point cloud with intensity values displayed in red and green for trunks and branches; COLMAP and NeRF models with color in RGB; 3DGS model with color in red.

Fig. 7 and 8 provide detailed views of different 3D point cloud models of Plot_1, including a detailed view from the same viewpoint, a single tree model, and a 10 cm-thick cross section of the tree trunk at 1.3 m. In Plot_1_Phone, the COLMAP model shows a dense distribution of ground points, whereas in NeRF and 3DGS models the ground points are sparser. NeRF exhibits noticeable gaps in the ground and errors in the positioning of ground points and tree trunks, but these issues are improved in the NeRF model of Plot_1_UAV. For single-tree models, the UAV images produce better results compared to the Phone images, evident in fewer noise points in the canopy and higher density of trunk points. In the cross-sectional profiles of tree trunks, Lidar and NeRF models exhibit relatively smooth hollow circles, COLMAP model shows erroneous overlapping positions that increase the trunk diameter, and 3DGS model forms a solid circle with a smaller diameter.

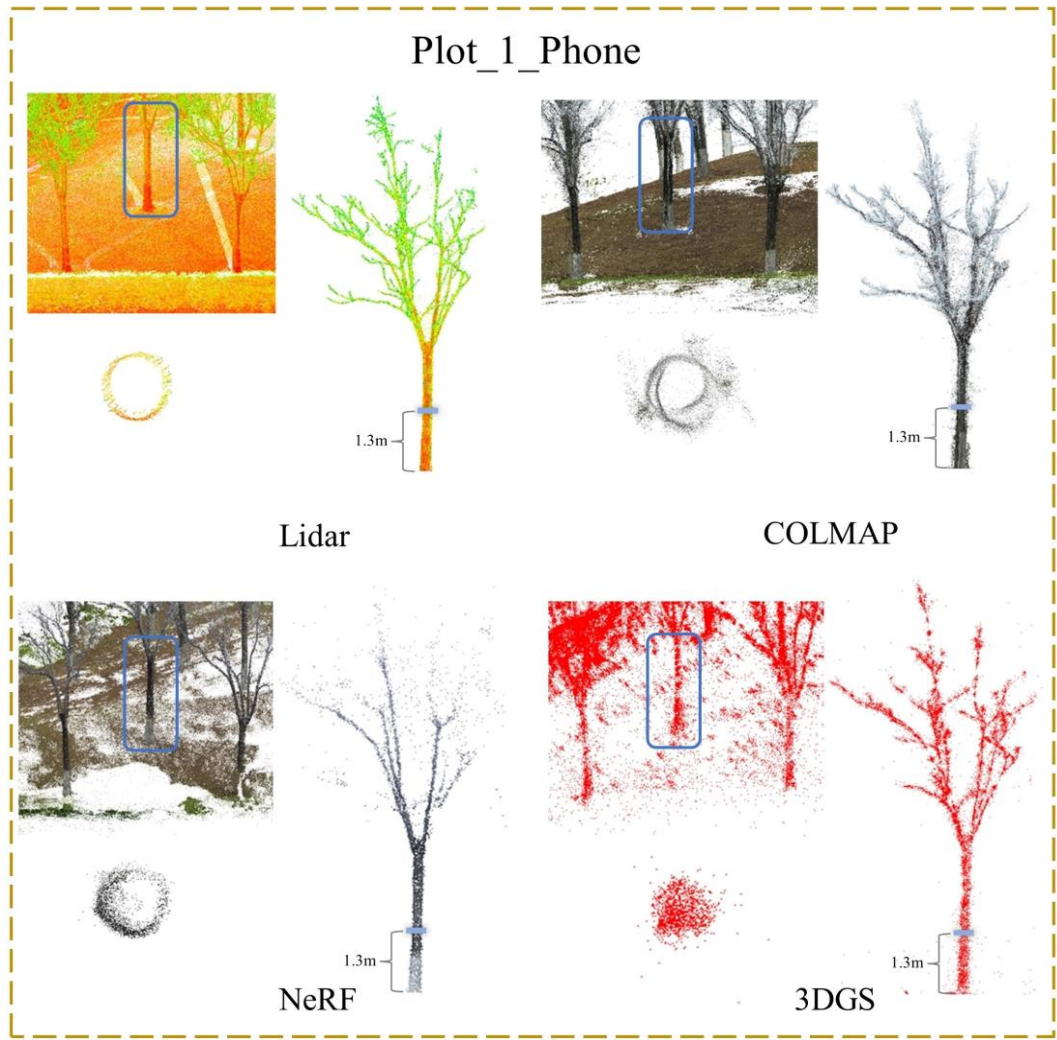

**Fig. 7.** Detailed comparison of the reconstruction results from Plot_1_Phone. This includes a more detailed view of the same view scene from the Lidar (TLS), COLMAP, NeRF, and 3DGS models, a single tree model extracted from the point cloud (blue box), and a cross-sectional profile of the tree trunk at 1.3m (the trunk is marked in blue).

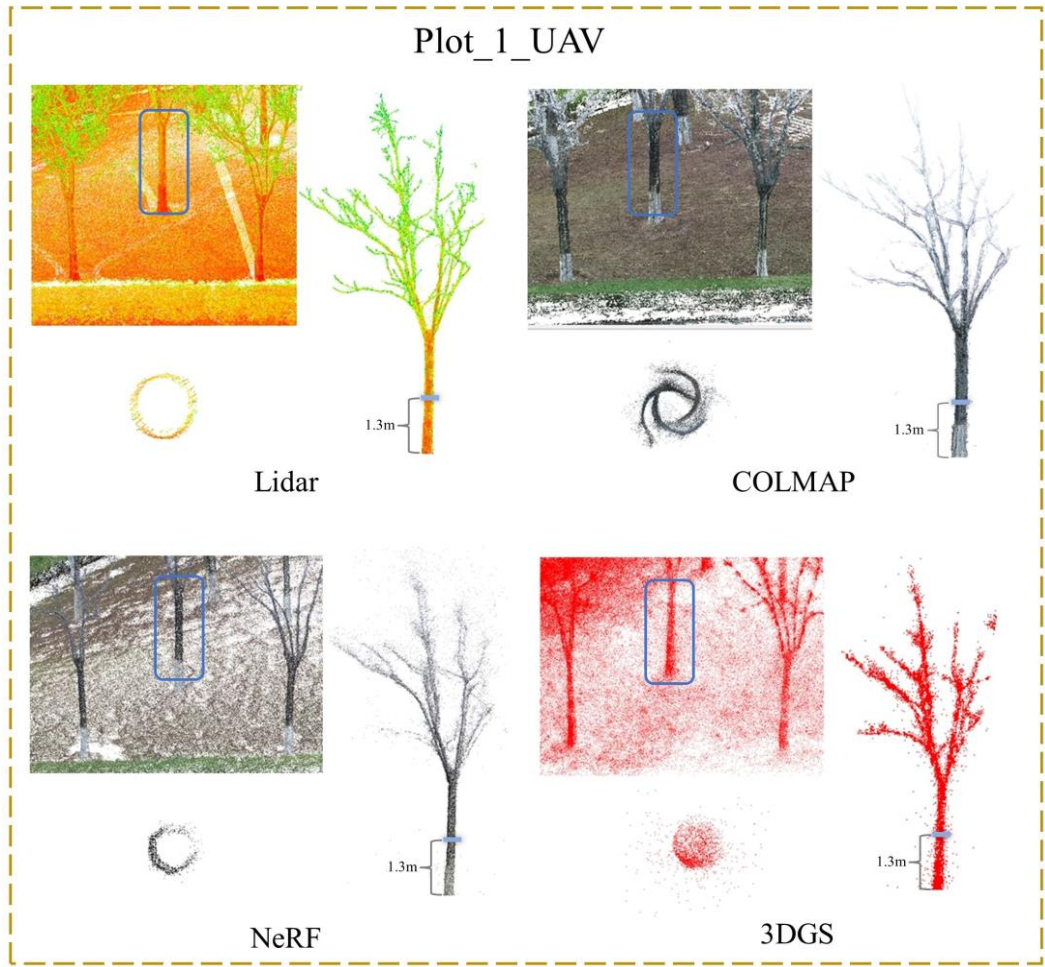

**Fig. 8.** Detailed comparison of Plot_1_UAV reconstruction results. This includes a more detailed view of the same view scene from the Lidar (TLS), COLMAP, NeRF, and 3DGS models, a single tree model extracted from the point cloud (blue box), and a cross-sectional profile of the tree trunk at 1.3m(the trunk is marked in blue).

Fig. 9 displays the 3D reconstruction results of Plot_2 also from two different perspectives. The results indicate that the NeRF model is the best, achieving photorealistic color and minimal noise. Although the COLMAP model has the highest point density, there are missing areas in the canopy and some tree trunks, with significant noise on the tree trunks. The 3DGS model is the sparsest, with more missing areas in the canopy and tree trunks compared to COLMAP and NeRF.

Fig. 10 shows more details of the four different models for Plot_2, including a bird's eye view of the model, a detailed view for a local region, a single tree model, and a 10 cm-thickness cross section of the tree trunk at 1.3 m. From the aerial view, it is evident that the canopy of some trees in the middle of the Lidar model is missing. This is partly due to the tall trees in Plot_2; the terrestrial laser scanner has limited scanning angles (100 degrees) from the ground, making it difficult to capture the higher parts of the canopy. Additionally, the dense leaf occlusion in the canopy limits the laser beam's penetration capability. The other three models all reconstruct the tree canopy, but the NeRF model has the highest canopy density and best quality. In terms of tree model details, the COLMAP tree model shows erroneous multiple trees at the same location and has more noise on the tree trunks, with misalignment in the trunk cross-sections. The NeRF tree model is closer to the Lidar model, reconstructing more tree trunks compared to COLMAP and 3DGS, with smoother trunk cross-sections. The 3DGS tree model has fewer and sparser trunk point clouds, which will affect the accuracy of trunk diameter extraction.

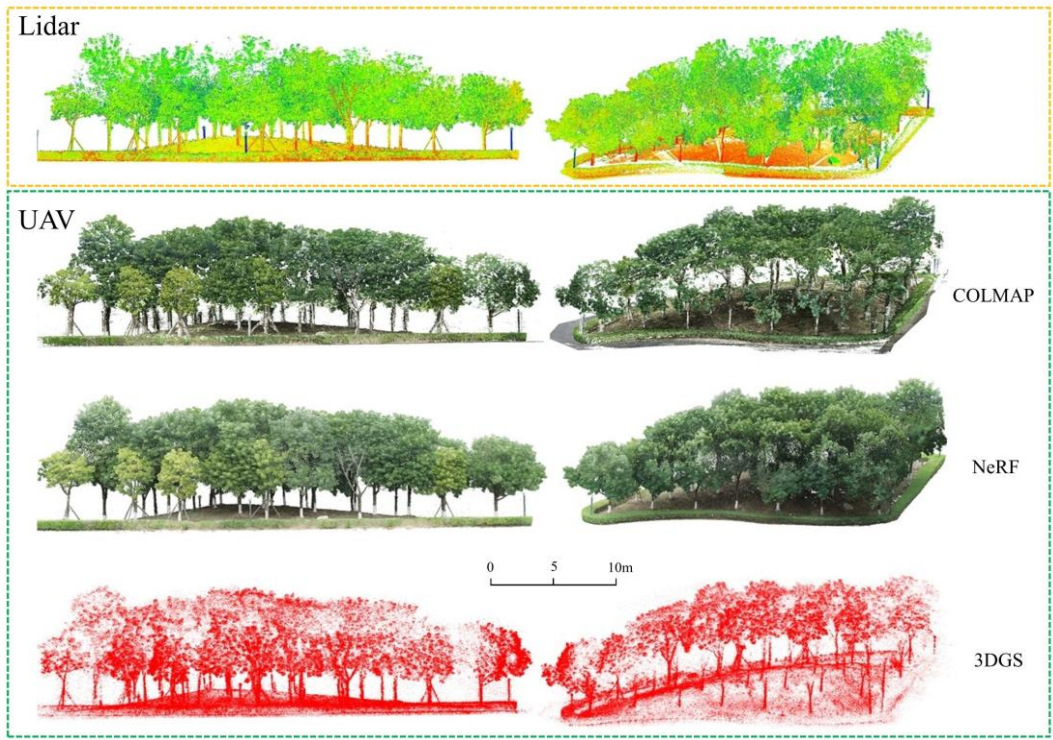

**Fig. 9.** Overall comparison of the reconstruction results for Plot_2 viewing from two perspectives. TLS Lidar point cloud with intensity values displayed in red for trunks and branches and green for leaves; COLMAP and NeRF models with color in RGB; 3DGS model with color in red.

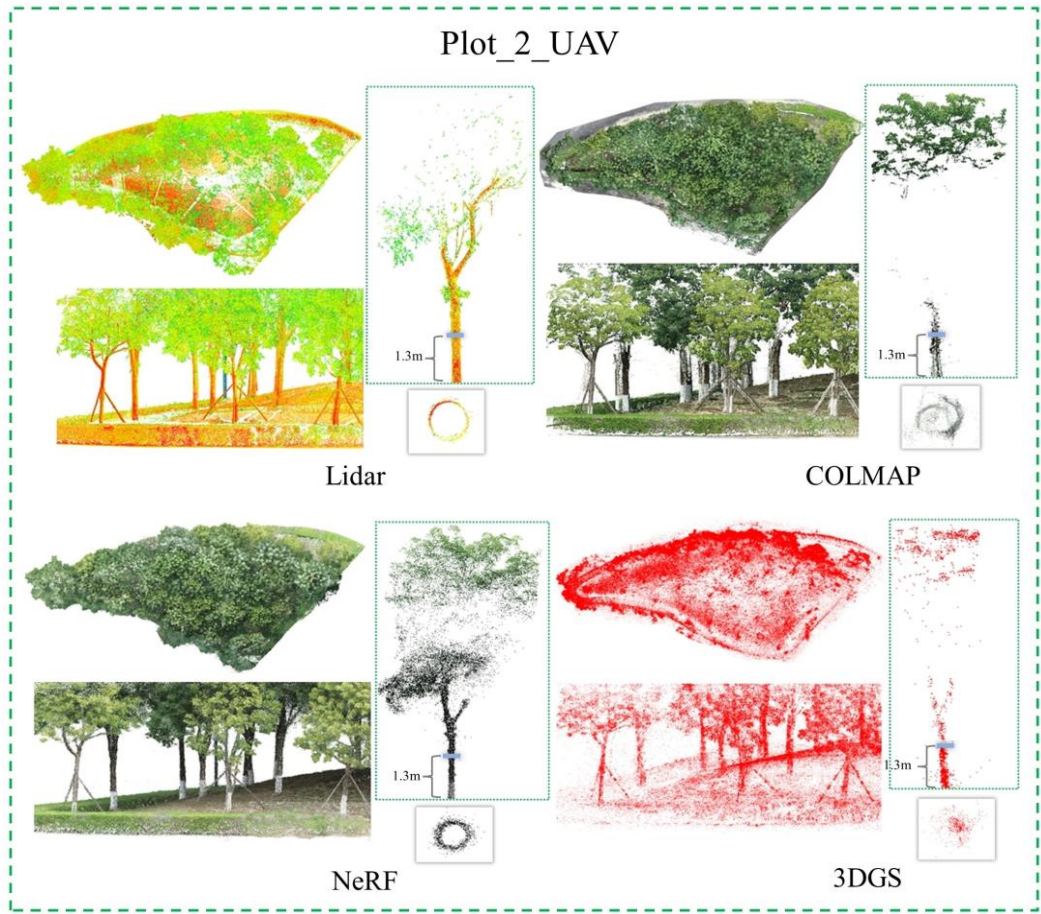

**Fig. 10.** Detailed comparison of the reconstruction results in Plot_2. This includes a top view of the Lidar (TLS), COLMAP, NeRF and 3DGS models, a more detailed view of the scene part, a single tree model extracted from the point cloud, and a cross-sectional profile of the tree trunk at 1.3m (the trunk is marked in blue).

*3.3 Extraction of Tree Parameters from stand plot Point Cloud*

We used LiDAR360 (https://www.lidar360.com/) to extract tree structural parameters from the Lidar (TLS), COLMAP, NeRF, and 3DGS point cloud models. The same procedures and parameters were used to process these point cloud in the software. We performed preprocessing tasks including point cloud filtering and normalization, and utilized single-tree segmentation tool to extract individual trees, and then determined the tree heights and DBHs in each plot. The metrics derived from the Lidar point cloud model served as a reference for comparison with those obtained from the three visual reconstruction models. The results are shown in Fig. 11.

The red numbers in the figure represent the actual number of trees measured in Plot_1 and Plot_2, which are 45 and 33 trees, respectively. The black numbers in the columns represent the number of trees obtained through single-tree segmentation based on the four point cloud models. In Plot_1, the number of trees in all four-point cloud models match the actual count. In Plot_2, the number of trees in the COLMAP model is 38, which is 5 more than the actual number, while the other models match the real count.

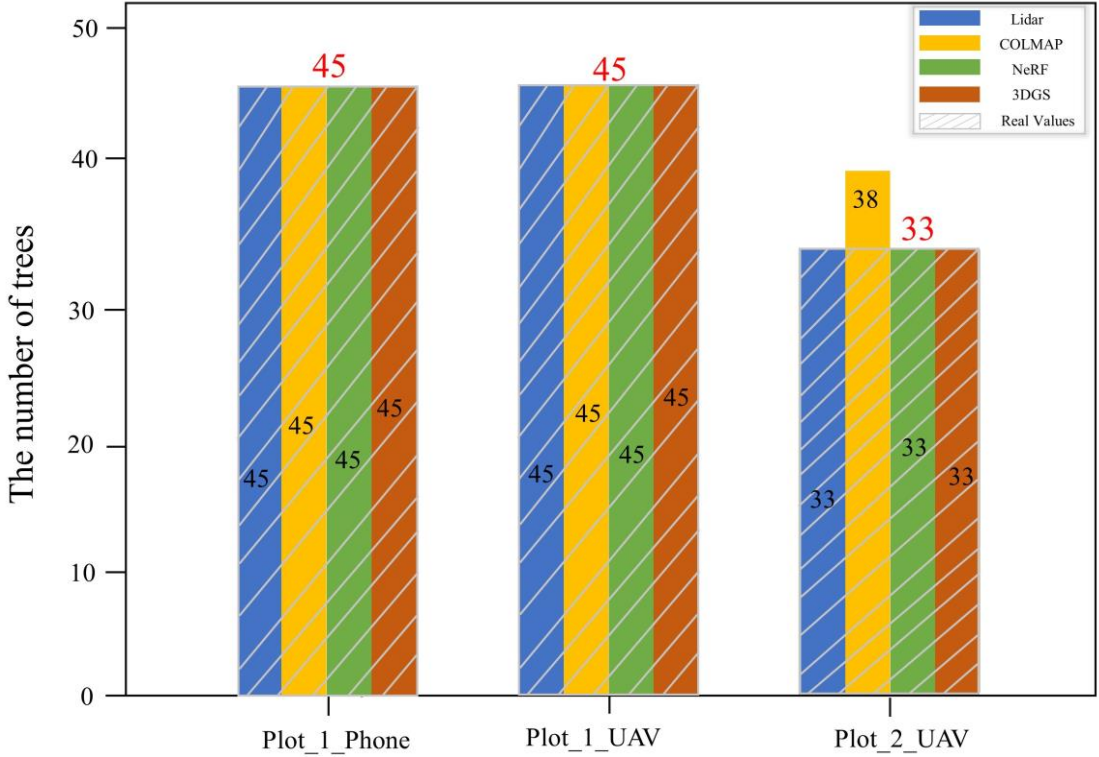

**Fig. 11.** Comparison of the number of individual trees extracted from the reconstruction results of Plot_1 and Plot_2. The red numbers in the columns represent the actual number of trees measured. The blue bars indicate the number of trees extracted from the Lidar (TLS) model, the yellow bars represent the COLMAP model, and the green bars show the 3DGS model, and the brown bars denote the NeRF model. The black numbers on top of the bars indicate the number of trees extracted from each corresponding model.

Based on the individual tree segmentation, structural parameters such as TH and DBH of each tree were extracted next. The coefficient of determination ($R^2$) and root mean square error ($RMSE$) were used as evaluation metrics for the structural parameters. The $R^2$ value assesses how well the regression model fits the reference data, with a value closer to 1 indicating a higher correlation between the data and a more accurate description of the variations. $RMSE$ represents the average deviation between the true values of the structural parameters and the observed values.

$$R^2 = 1 - \frac{\sum_{i=1}^{m}(y_i - \hat{y}_i)^2}{\sum_{i=1}^{m}(y_i - \bar{y})^2} \qquad (3)$$

$$RMSE = \sqrt{\frac{1}{m}\sum_{i=1}^{m}(y_i - \hat{y}_i)^2} \quad (4)$$

where $\hat{y}_i$ and $y_i$ represent the predicted and observed values of the samples, respectively, $\bar{y}$ denotes the mean of the observed values, and m is the number of samples.

Fig. 12 and 13 display the results of linear fitting between the TH and DBH extracted from three models and the Lidar reference values for Plot_1. For three height TH, Phone_COLMAP and Phone_3DGS show good fitting with Lidar, with *RMSE* values of 0.9215 and 0.8769, and *RMSE* values of 0.4592 m and 0.5006 m, respectively. Phone_NeRF suffers from misalignment of ground and tree trunk points, which affects the accuracy of TH extraction, resulting in an $R^2$ of 0.7639 and an *RMSE* of 0.7116 m. The three models reconstructed from UAV imagery improve the accuracy of TH extraction, with $R^2$ values of 0.9375, 0.8960, and 0.9526 for COLMAP, 3DGS, and NeRF, respectively, representing improvements of 1.60%, 1.91%, and 18.87% compared to the Phone models. *RMSE* values also decrease to 0.4333 m, 0.4014 m, and 0.2955 m, respectively.

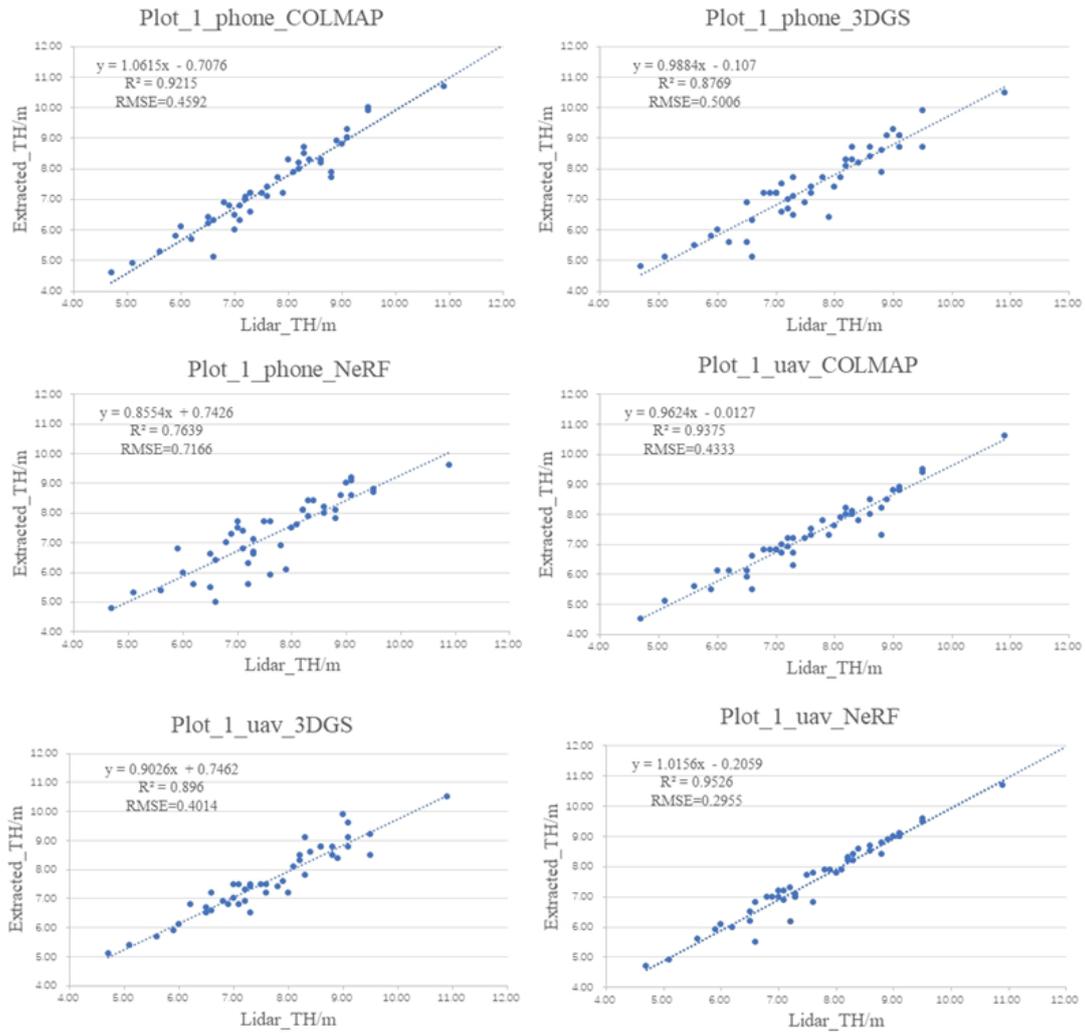

**Fig. 12.** Linear fitting results of TH extracted from Plot_1_Phone and Plot_1_UAV models compared to those derived from Lidar (TLS) reference values.

In the DBH extractions for Plot_1, the three point cloud models from Phone do not fit the DBH well, with $R^2$ values all below 0.5, where Phone_NeRF has the best fit at 0.4388. The UAV_COLMAP and UAV_NeRF models are closer to Lidar values compared to those of Phone models, with $R^2$ values of 0.8825 and 0.8381, representing increases of 62.41% and 39.93% respectively. The *RMSE* values have also decreased to 0.9928cm and 1.1741cm. The Phone_3DGS and UAV_3DGS models show much smaller DBH values than the Lidar reference due to the sparse nature of the tree trunks in these models.

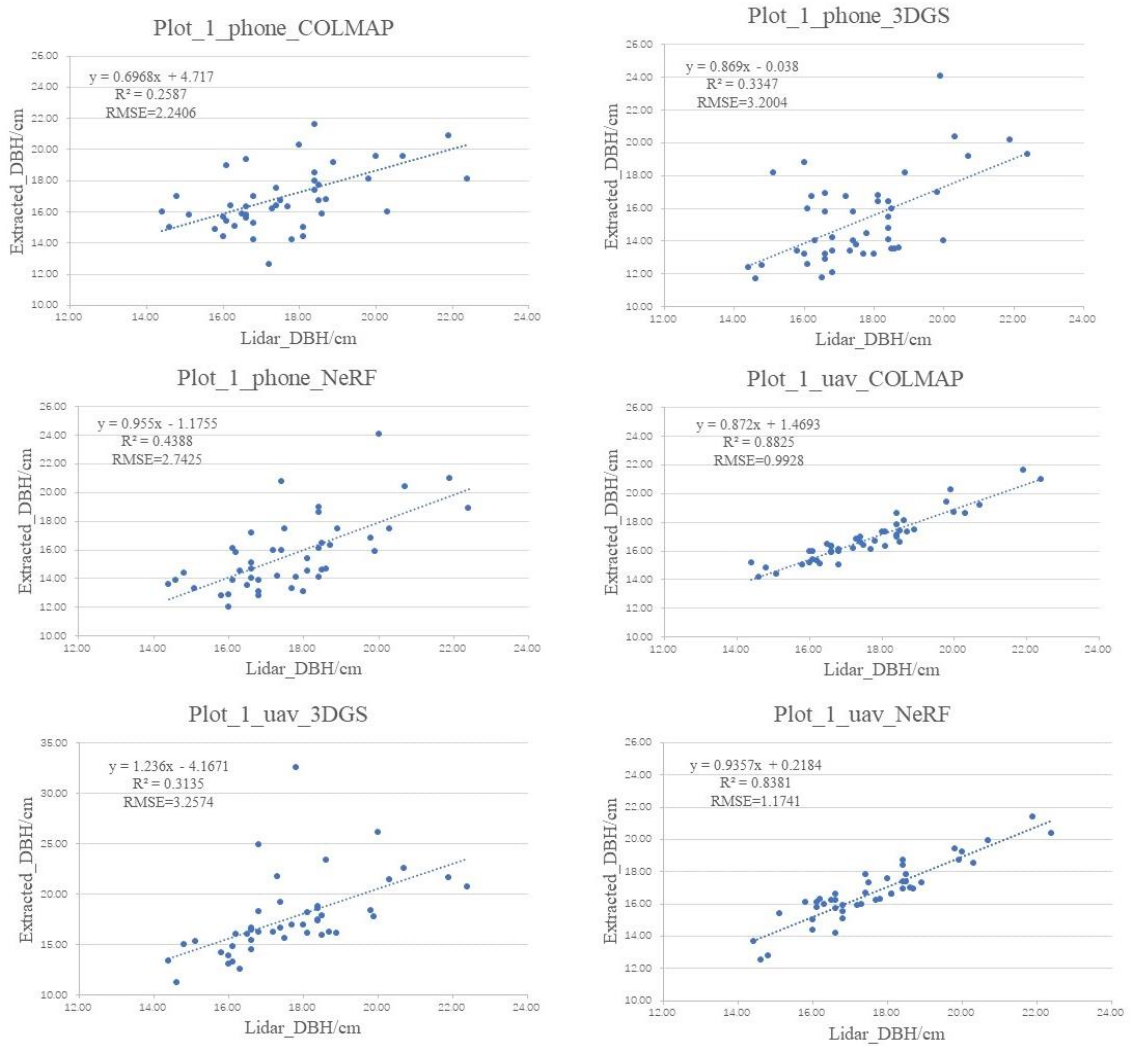

**Fig. 13.** Linear fitting results of DBH extracted from Plot_1_Phone and Plot_1_UAV models compared to those derived from Lidar (TLS) reference values.

Fig. 14 and Fig. 15 show the linear fitting results of TH and DBH extracted from models compared to those derived from Lidar reference values for Plot_2. In the TH extractions, the Plot_2_Lidar model is missing canopy information for some trees in the middle section, making these trees unsuitable for comparison. Thus, after excluding these trees from the Lidar_TH values, comparisons were made with TH values extracted from COLMAP, 3DGS, and NeRF models. The results indicate that COLMAP, 3DGS, and NeRF models achieved high accuracy in fitting Lidar values. COLMAP's $R^2$ reached 0.8792, with an *RMSE* of 0.3252m. 3DGS and NeRF models have $R^2$ values of 0.9289 and 0.9281, with *RMSE* of 0.4751m and 0.3151m, respectively, indicating that their extracted TH values are closer to Lidar TH values when compared to COLMAP.

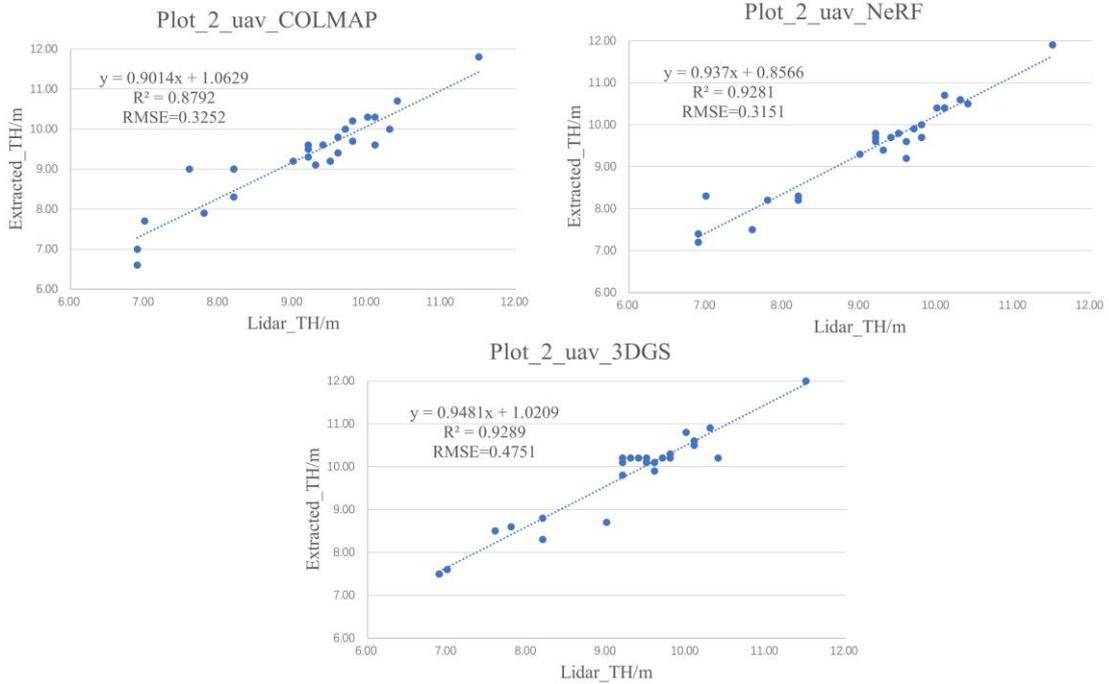

**Fig. 14.** Linear fitting results of TH extracted from Plot_2_UAV model compared to those derived from Lidar (TLS) reference values.

Due to the sparse distribution of tree trunk points in the Plot_2_UAV_3DGS model, DBH fitting could not be performed, and some reconstructed tree trunks did not reach the required height for DBH (1.3m). Therefore, DBH values could not be successfully extracted. The DBH values obtained from COLMAP and NeRF are relatively close to the Plot_2_Lidar reference values, with $R^2$ values of 0.8747 and 0.8648, and *RMSE* values of 1.6067 cm and 1.6327 cm, respectively.

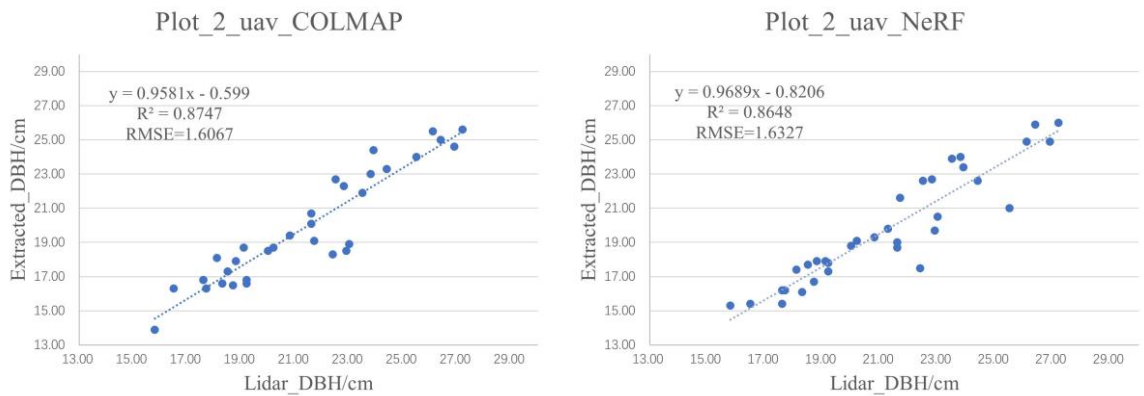

**Fig. 15.** Linear fitting results of DBH extracted from the Plot_2_UAV model compared with those derived from Lidar (TLS) reference values.

## 4. Discussion

In this paper, we applied three image-based 3D reconstruction techniques in forest stands, including a traditional photogrammetry pipeline (SfM+MVS in COLMAP) and novel view synthesis-based methods (NeRF and 3DGS). By comparing the dense point cloud models generated by these three methods with the reference TLS point clouds, we analyzed the reconstruction efficiency and point cloud model quality of different methods for reconstructing forest stands.

The difficulty in successfully reconstructing multi-view 3D models of forest stands typically lies in the fact that trees of the same species have similar texture structures, making it challenging for algorithms to distinguish （detect and match）tree features. Additionally, due to the occlusion of tree branches and canopies, the views could change dramatically even between two adjacent images. The first step for photogrammetry, NeRF, and 3DGS is to obtain accurate camera poses and calibrated images through SfM. If the results of SfM are inaccurate, it will affect the subsequent dense reconstruction outcomes, for example, duplicated (phantom) trunks can appear at the same location with a small offset. In the COLMAP pipeline, the steps of SfM include feature extraction, feature matching, and triangulation of points. During our SfM experiments, we found that the results of feature matching are related to the completeness of the camera pose estimation. By default, COLMAP uses Exhaustive matching (where all images are matched pairwise). Using this method for matching the images of Plot_1_UAV and Plot_2_UAV, SfM was only able to successfully solve for the poses of 173 images (out of total input of 268 images) and 268 images (input of 322 images), respectively, along with generating a sparse point cloud for parts of the scene. However, COLMAP also supports other matching methods, such as spatial matching (which requires images to have positional information, such as GPS data). Considering that the UAV images of Plot_1 and Plot_2 contain positioning information, using spatial matching allows for the successful retrieval of the complete camera pose information for all input images (Plot_1_UAV: 268 images, Plot_2_UAV: 322 images) and a more complete sparse point cloud. The results of downstream dense reconstruction (MVS, NVS) depend to large extent on the accuracy of the upstream SfM results, which is why spatially matched SfM results are used during the dense reconstruction phase. To reduce dependence on SfM, some studies have utilized optical flow and point trajectory principles for camera pose estimation (Smith et al., 2024). Others have integrated the camera pose estimation step into the 3DGS network training framework, simultaneously optimizing both 3DGS and camera poses (Yu et al., 2024). Additionally, the scene regression-based ACE0 (Brachmann et al., 2024) method and the general-purpose global SfM method GLOMAP (Pan et al., 2024) significantly improve operational efficiency and achieve SfM estimates comparable to COLMAP. More robust feature matching methods, such as LoFTR (Sun et al., 2021) and RCM (Lu and Du, 2024), can also replace the feature matching step in SfM to improve matching accuracy.

MVS utilizes the corresponding images and camera parameters obtained from sparse reconstruction to reconstruct a dense point cloud model through multi-view stereopsis. This process involves calculating depth information for each image and fusing depth maps. Typically, the time required to compute depth information occupies the majority of the MVS dense reconstruction time. Moreover, the time consumption increases with the size, complexity of the scene, and the number of images. In previous study on single tree reconstruction, the MVS dense reconstruction of a single tree (with 107-237 input images) usually took 50 to 100 minutes on a single 3090 GPU(Huang et al., 2024). However, in this study, despite using the more computationally efficient 4090 GPU, the reconstruction of forest plot scenes (with 268-322 input images) still took 450 to 700 minutes. The average processing time per input image increases from 0.43-0.83 minutes to 1.40-2.70 minutes, an approximate increase of about 3.3 times. In contrast, efficient deep learning networks like NeRF and 3DGS can significantly reduce the reconstruction time, usually completing within 20 minutes, and the time required appears to be independent of the scene size and the number of images.

By comparing the number of points, overall view and details of the point cloud models, we highlight the advantages and disadvantages of the three image-based reconstruction methods. The COLMAP method can generate models with the highest number of points, but it tends to introduce more noise in the trunk and canopy layers. Comparing the COLMAP models of Plot_1_Phone and Plot_1_UAV, the UAV provides more image data for the canopy, which reduces noise in the UAV_COLMAP model's canopy region. However, a significant amount of noise remains. In contrast, the NeRF model exhibits less noise in the canopy. When there is ground occlusion or limited viewpoints, such as in the case of Plot_1_Phone, the NeRF model exhibits missing or erroneous reconstructions in the ground regions. The 3DGS model contains the fewest points, with fewer than 2 million points overall, resulting in a sparse and low-quality 3DGS point cloud model that fails to represent real-world colors accurately. This indicates that the ability of 3DGS to generate dense point cloud models is inferior to that of COLMAP and NeRF. In the more complex scene of Plot_2, NeRF can produce point cloud models closest to those obtained from LiDAR, compared to COLMAP and 3DGS models. NeRF reconstructs better real-world colors, provides more detailed trunks, and captures more complete canopies. Meanwhile, the COLMAP model often shows intersecting trunks or multiple overlapping trees, demonstrating that NeRF is better suited to handling complex forest stand.

When selecting plots and collecting data, different types of plots and data acquisition methods were chosen with the aim of comparing and illustrating how these factors might affect the results. We selected two plots with different tree species and canopy morphologies: Plot_1 consisted of leafless trees with simple branch structures, resulting in more complete tree point cloud models. In contrast, Plot_2 had a higher tree density, which led to incomplete tree trunks in the middle of the plot. Additionally, the dense foliage in Plot_2 caused the inner branches of the canopy to be missing. For Plot_1, image data were collected using both a smartphone (capturing lower-resolution images from the ground only) and a UAV (capturing higher-resolution images from both ground and aerial perspectives). The UAV images, which provided more viewpoints, resulted in more complete tree point cloud models with fewer noise points in the canopy. Furthermore, models generated from higher-resolution images contained more detailed tree features. The quality of the models also significantly impacted the accuracy of subsequent individual tree structure parameter extraction.

When TLS is used to acquire 3D point cloud models of target objects solely from ground-level perspectives, particularly in forest scenes with tall trees and dense canopies, the canopy of the middle sections of trees often becomes incomplete, affecting the accuracy of TH extraction. To obtain a complete and accurate canopy, a combination of TLS and ALS is necessary, which inevitably increases workload and costs. In contrast, using UAV to capture images of tree canopies is simple, convenient, and cost-effective. Furthermore, image-based visual reconstruction methods can also effectively reconstruct tree canopies and accurately extract TH.

We performed individual tree segmentation based on different point cloud models and extracted TH and DBH as individual tree structural parameters for comparison. During the individual tree segmentation process, the number of trees in the Plot_2_COLMAP model exceeded the actual count. This could be attributed to the high similarity in texture features and partial occlusion among trees in this plot, which reduces the number of feature points and results in mismatches, causing a single tree to be represented by multiple duplicated models. Therefore, the traditional photogrammetry algorithm (SfM+MVS) sometimes is unable to handle more complex forest plot scenes. In terms of TH parameters, all three methods provide high accuracy, with estimates from the NeRF model generally outperforming those from the COLMAP and 3DGS models. In terms of DBH estimation, the models generated from lower-resolution smartphone images show poorer accuracy, while the models reconstructed from higher-resolution UAV images demonstrate higher precision. Among them, the COLMAP model provides better DBH estimates; however, in the extracted individual COLMAP tree models, some phantom tree trunks often intersect each other, which may cause the estimated DBH values to be larger than the actual ones. The NeRF method has slightly lower estimation accuracy compared to the photogrammetry approach and tends to underestimate the DBH values. The 3DGS method yields the least accurate DBH estimates, as its reconstructed points are relatively sparse and trunk points tend to clustering together, leading to DBH estimates that are significantly lower than the reference LiDAR DBH values.

Through this study, we have gained a deeper understanding of the practical applications of NeRF and 3DGS methods in forest scenes and the properties of the dense point cloud data generated by these methods, providing answers to the questions raised in the introduction earlier. At the same time, we outline prospects for future research: how to improve the accuracy of sparse point clouds and camera pose parameters obtained from upstream SfM, potentially by using more robust feature matching techniques and alternative SfM solutions, and by enhancing feature matching success rates through image enhancement methods; how to enhance the ability of NeRF and 3DGS to generate dense point clouds. We anticipate that the emergence of more powerful software tools and well-designed strategies will make efficient and higher-precision 3D reconstruction of forest scenes possible.

## 5. Conclusions

This study applied NVS (NeRF and 3DGS) techniques to the 3D dense reconstruction of forest stand. Using two forest plots with different tree structure and morphology (one leaf-on and one leaf-off) as examples, the feasibility of these techniques was validated. Specifically, dense point cloud models were generated using NVS techniques based on sequential images captured from different devices and viewpoints, and were compared with traditional photogrammetric methods. A comprehensive evaluation of the practical application of different NVS methods in forest stand was conducted in terms of processing efficiency, point cloud model quality, and the accuracy of tree parameters extraction. The results indicated that:

1) The new view synthesis methods (NeRF and 3DGS) achieve significantly higher efficiency in dense reconstruction compared to traditional photogrammetry methods.

2) The 3DGS method's capability to generate dense 3D point clouds is inferior to that of NeRF and traditional photogrammetry methods, with 3DGS models often exhibiting sparser point densities and being inadequate for single-tree diameter estimation.

3) For complex forest stand with dense foliage and high tree density, NeRF provides superior reconstruction quality, while photogrammetry methods tend to produce poorer results, including issues such as tree trunk overlap and multiple tree duplications.

4) All three methods achieve high accuracy in extracting single-tree height parameters, with NeRF providing the highest precision for tree height. Traditional photogrammetry methods offer better accuracy in diameter estimation compared to NeRF and 3DGS.

5) The quality of images (resolution and quantity) and the completeness of viewpoints significantly impact the quality of the reconstruction results and the accuracy of tree structure parameter extraction.

**CRediT authorship contribution statement**

**Guoji Tian:** Writing – original draft, Writing – review & editing, Visualization, Validation, Software, Methodology, Formal analysis, Data curation. **Hongyu Huang:** Writing – review & editing, Methodology, Conceptualization, Formal analysis, Software, Supervision, Project administration. **Chongcheng Chen:** Supervision, Project administration.

**Funding:**

The research receives no funding.

**Data availability**

Data will be made available on request.

**Acknowledgments**

We would like to thank Luyao Yang and Hangui Wang for their assistance in collecting and processing terrestrial laser scanning data.

**References：**